\documentclass{IEEEcsmag_arxiv}

\usepackage[colorlinks,urlcolor=blue,linkcolor=blue,citecolor=blue]{hyperref}

\usepackage{upmath}
\usepackage{booktabs}
\usepackage{multirow}
\usepackage[table,xcdraw]{xcolor}
\usepackage{adjustbox}

\begin{document}


\title{AI Maintenance:\\ A Robustness Perspective}

\author{Pin-Yu Chen and Payel Das}
\affil{IBM Research \\ pin-yu.chen@ibm.com and daspa@us.ibm.com}


\begin{abstract}
With the advancements in machine learning (ML) methods and compute resources, artificial intelligence (AI) empowered systems are becoming a prevailing technology. However, current AI technology such as deep learning is not flawless. The significantly increased model complexity and data scale incur intensified challenges when lacking trustworthiness and transparency, which could create new risks and negative impacts. In this paper, we carve out AI maintenance from the robustness perspective. We start by introducing some highlighted robustness challenges in the AI lifecycle and motivating AI maintenance by making analogies to car maintenance. We then propose an AI model inspection framework to detect and mitigate robustness risks. We also draw inspiration from vehicle autonomy to define the levels of AI robustness automation. Our proposal for AI maintenance facilitates robustness assessment, status tracking, risk scanning, model hardening, and regulation throughout the AI lifecycle, which is an essential milestone toward building sustainable and trustworthy AI ecosystems.
\end{abstract}

\maketitle

\section{Introduction}
Just like the indispensable role of cars in the modern world, AI-empowered technology, and ML-based systems and algorithms are bringing revolutionary changes and far-reaching impacts on our life, society, and environment, if not happening already. As AI models are perceived as a new ``vehicle'' to a better future, this article aims to stress the importance of formalizing and practicing \textit{AI maintenance} from the robustness perspective,
by drawing analogies in the model development and deployment between car and AI. Towards achieving trustworthiness and sustainability for AI, 
this article is motivated by the following question: 
\textit{Cars require regular inspection, maintenance, and continuous status monitoring, why should AI technology be any different?}

Robustness in AI often entails multiple meanings depending on the context and use cases. In this article, we study robustness from the perspective of the generalization capability of an AI model in \textit{adversarial} and \textit{unseen} scenarios. In general, the performance of an AI model is evaluated in the \textit{average case}, by comparing the model predictions on a set of data samples to their ground-truth labels and then using the average prediction result as a performance metric, such as the top-1 classification accuracy measuring the fraction of correct model prediction on the most-likely (top-1) class over a dataset. In contrast, the adversarial scenario evaluates the model performance in the \textit{worst case} among all possible and plausible changes (often pre-specified) to the data and AI model, by assuming a virtual adversary is in place.  
Moreover, the unseen scenario evaluates the model performance on new data samples that are drawn from a different data distribution than the seen data samples during training (but not necessarily the worst-case distribution), possibly caused by natural data/label shifts, and real-world observational noises, among others.

The rationale for studying AI maintenance from the robustness viewpoint is motivated by the rapidly intensified demand for inspecting and preventing failure modes for AI models, in order to understand the limitations and prepare AI technology for the real world against malicious attempts and contiguous data changes. 
According to a recent Gartner report\footnote{\url{https://www.gartner.com/smarterwithgartner/gartner-top-10-strategic-technology-trends-for-2020}}, 30\% of cyberattacks by 2022 will involve data poisoning, model theft or adversarial examples (see \cite{chen2022holistic} for an overview of these new risks centered on machine learning). However, the industry seems underprepared. In a survey of 28 organizations spanning small and large organizations, 25 organizations did not know how to secure their AI/ML systems \cite{kumar2020adversarial}. Unlike car insurances that cover damage and liability, 
the risk of lacking robustness in AI models can be further amplified if cyber insurance providers impose stringent requirements when the root cause is related to AI failure modes\footnote{\url{https://hbr.org/2020/04/the-case-for-ai-insurance}}. Moreover, AI maintenance is closely related to action plans for enhancing trustworthiness in safety-related ML applications, such as fulfilling the milestones and objectives defined in the roadmap of the European Union Aviation Safety Agency (EASA)\footnote{\url{https://www.easa.europa.eu/newsroom-and-events/news/easa-releases-its-concept-paper-first-usable-guidance-level-1-machine-0}}, the AI/ML Software as a Medical Device Action Plan defined by U.S. Food \& Drug Administration\footnote{\url{https://www.fda.gov/medical-devices/software-medical-device-samd/artificial-intelligence-and-machine-learning-software-medical-device}}, and the NIST AI Risk Management Framework\footnote{\url{https://www.nist.gov/itl/ai-risk-management-framework}}.

\begin{figure*}[t]
    \centering
    \includegraphics[width=1\textwidth]{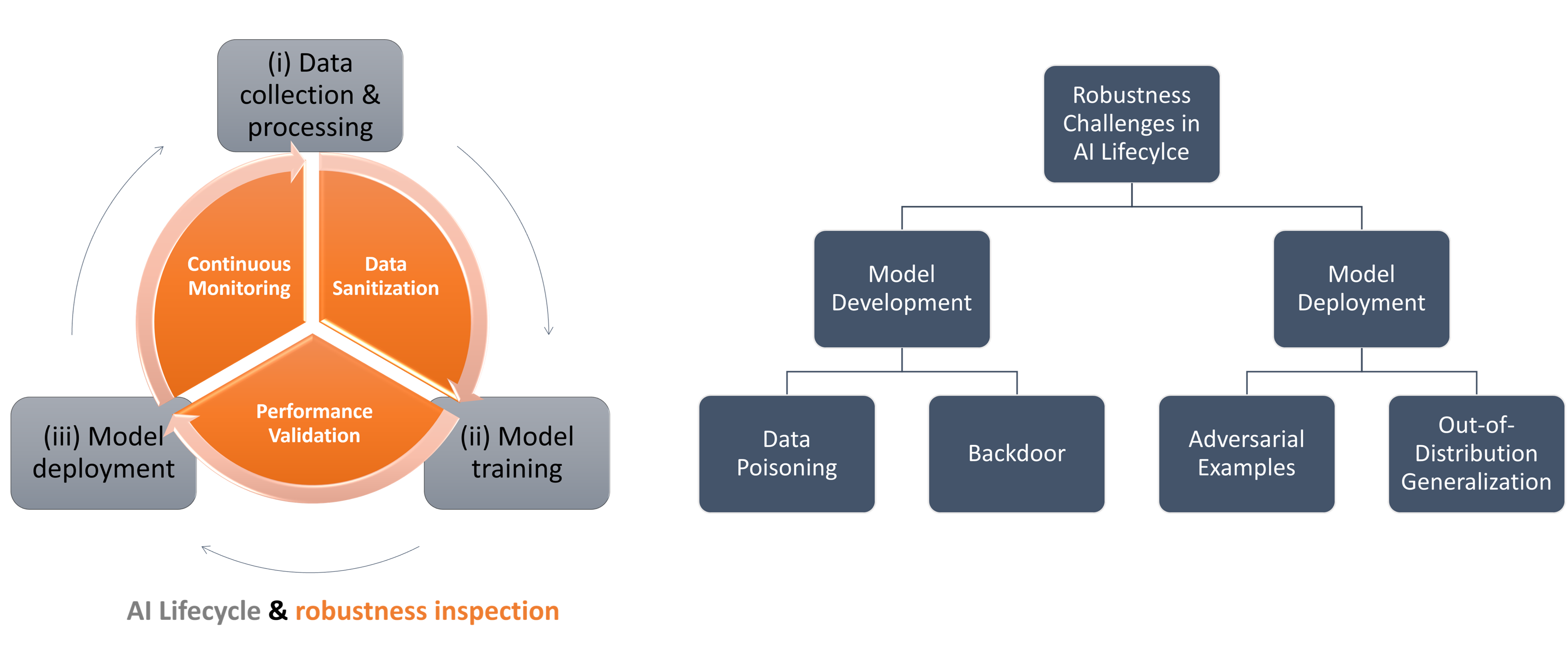}
    \caption{\textbf{Left:} Schematic illustration of robustness inspection pipeline (data sanitization, performance validation, and continuous monitoring) in the AI lifecycle consisting of three major states: data collection and processing, model training, and model deployment. The model development phase includes data collection and processing and model training.
    \textbf{Right:} Highlighted robustness challenges in the AI lifecycle. In the model development phase, the robustness challenges assume the training data are subject to manipulation prior to model training. In the model deployment phase, the robustness challenges have no access to the training data but may assume some knowledge of the deployed model such as the model architecture and the associated model parameters. Based on the categorization of states in the AI lifecycle, the chart can be extended to incorporate other robustness challenges and other trustworthiness dimensions such as safety, privacy, etc.}
    \label{fig:lifecycle_robustness}
\end{figure*}

To gain insights into AI maintenance, this article first introduces major robustness challenges in the AI lifecycle for model development and deployment. Then, we make analogies of the commonality between car and AI maintenance. Finally, we propose the conceptual framework named ``AI model inspector'' for holistic robustness inspection and enhancement. Similar to the definitions of driving automation for vehicle autonomy, we define six levels of AI robustness towards facilitating qualitative and quantitative assessment of AI technology throughout the entire lifecycle.



\section{Robustness Challenges in AI Lifecycle }

\textbf{Figure \ref{fig:lifecycle_robustness}} provides an overview of robustness inspection pipeline in the AI lifecycle (left panel) and the highlighted robustness challenges (right panel). The AI lifecycle is recurring between two phases: model development and deployment. The model development phase consists of two states: (i) data collection and processing, and (ii) model training. Data collection and processing include typical data operations such as data acquisition and labeling, feature normalization, filtering, anonymization, and data augmentation.
Model training involves machine learning model selection, algorithm development, system design, and optimization. Between states (i) and (ii), data sanitization inspects the data fidelity and performs mitigation steps (e.g., deleting problematic data samples or correcting mislabeled samples) prior to model training.
After model development, the AI lifecycle enters the state of (iii) model deployment, in which the trainable model parameters are frozen for use.
Between (ii) and (iii), performance validation inspects and reduces the gap between model training and deployment. If the deployed model undergoes significant performance degradation, possibly due to naturally occurring data shifts or malicious attempts, the AI lifecycle will re-enter the model development phase to collect new data or update the model. Between (iii) and (i), continuous monitoring inspects the performance status of the currently deployed model and gives a notice upon observing significant performance degradation or detecting anomalous events.

There are different types of robustness challenges in the model development and deployment phases that can lead to model misbehavior and degraded performance, varied by their objectives, feasible actions on intervening in the AI model, and knowledge about the AI model. In the adversarial scenario, the robustness challenges can be related to a ``threat model'' specifying what an attacker can know and do to compromise the AI model. In the unseen scenario, the robustness challenges are associated with the domain generalization capability between the development and deployment phases. Figure \ref{fig:lifecycle_robustness} (right panel)
lists two highlighted robustness challenges for each phase, which are detailed as follows.

\subsection{Robustness challenges in development phase}

\textit{Data poisoning} concerns the model performance when trained on noisy data. The source of noise may come from imperfect data collection and processing such as incorrect data annotation, data bias and imbalance, and context-irrelevant spurious features. The noise may also be intentionally introduced to the training data by adding a set of poisoned data samples for the purpose of undermining the model performance in the deployment phase. For example, making the target model has low classification errors in development but high classification errors in deployment.
Such intentional data poisoning attacks usually assume the ability to manipulate the training data and have access or some partial knowledge about the model details and training procedure \cite{jagielski2018manipulating}.  

\textit{Backdoor} is a Trojan attack targeting machine learning \cite{BadNet_Access}. It works by injecting some pattern (a trigger) with modified labels to a subset of training data. 
Due to the memorization effect of state-of-the-art machine learning models such as neural networks, models trained on the tampered dataset will contain a backdoor.
In the deployment phase, backdoored models will allow an attacker to gain control of the model output in the presence of the designated trigger, regardless of the actual content of the data input. However, in the absence of the trigger, the backdoored model will behave like a normal model trained on the untampered training dataset. Therefore, backdoor attacks are stealthy because the tampered model will not misbehave if the backdoor is inactivated. This challenge can be amplified in distributed and decentralized machine learning paradigms involving multiple parties exchanging limited information about their local private data, such as federated learning \cite{xie2020dba}.

\subsection{Robustness challenges in deployment phase}

The deployment phase takes a fully-tuned model in the development phase and freezes the model for subsequent data inference tasks. A deployed model is called a white-box model if its details are transparent to a user (e.g., releasing a deep learning model with its model architecture and pre-trained weights). Otherwise, if model details are unknown (or partially known) to a user, it is called a black-box (gray-box) model, such as a prediction application programming interface (API) or proprietary software that only gives model prediction results and does not reveal other details. For robustness assessment, the white-box mode enables full-stack system debugging and internal penetration testing, while the black-box mode allows practical vulnerability and information leakage analysis based on user access.

\textit{Adversarial examples} are carefully crafted data samples that cause prediction evasion when compared to the original unmodified data samples \cite{goodfellow2014explaining}. The easiness in prediction evasion reflects the model sensitivity against small changes in data inputs, such as a human-imperceptible additive perturbation. The robustness challenges of adversarial examples are often associated with safety-critical and security-related AI applications, such as autonomous driving cars, identification and recognition, and malware detection because their existence can be interpreted as counter-examples that violate the required robustness constraints.
In the black-box setting, adversarial examples can be generated by iteratively modifying a data input based only on the model's prediction output \cite{CPY17zoo}.

\textit{Out-of-distribution generalization} refers to the characterization of model performance when the input data samples undergo certain semantic-preserving transformations that deviate from the seen data distribution during model training. 
In contrast, in-distribution generalization refers to the model performance on data samples or instances drawn from the same distribution as the training data or environments.
The quest for out-of-distribution generalization is motivated by retaining robust predictions against natural variations (their effect can be either observable or hidden). The examples include distributional shifts between development and deployment phases,
data/label drifts in online data streaming, common corruptions caused by measurement/device errors, and data-invariant operations made by image rotation or scaling. An ideal model in deployment should generalize well or has the ability to quickly recognize and adapt to unseen data samples that are out-of-distribution yet share similar contexts to the in-distribution data seen during training.

\begin{table*}[t]
\caption{Analogies between car and AI models for maintenance and robustness divided into four categories. }
\label{tab: car_AI}
\centering
\begin{adjustbox}{max width=0.95\textwidth}
\begin{tabular}{l|l|l}
\toprule
\textbf{Category}                                                                                                  & \textbf{Car}                         & \textbf{AI}                       \\ \hline
\multirow{4}{*}{\begin{tabular}[c]{@{}l@{}}Model descriptions \\ and performance \\ characterization\end{tabular}} & user manual                          & model specification               \\ \cline{2-3} 
                                                                                                                   & automobile parts                     & machine learning modules          \\ \cline{2-3} 
                                                                                                                   & warrant                              & robustness checkpoints            \\ \cline{2-3} 
                                                                                                                   & transmission efficiency              & memory/data/power efficiency      \\ \hline
\multirow{4}{*}{\begin{tabular}[c]{@{}l@{}}Systematic inspection \\ and monitoring\end{tabular}}                   & collision test \& safety report      & internal robustness assessment    \\ \cline{2-3} 
                                                                                                                   & mechanical and electrical inspection & penetration testing and debugging \\ \cline{2-3} 
                                                                                                                   & problematic status warning            & operational errors                \\ \cline{2-3} 
                                                                                                                   & health state monitoring              & model behavior tracking           \\ \hline
\multirow{4}{*}{Fix and update}                                                                                    & repair                               & model fix and update              \\ \cline{2-3} 
                                                                                                                   & wheel alignment                      & model calibration                 \\ \cline{2-3} 
                                                                                                                   & winter tire                          & model hardening                   \\ \cline{2-3} 
                                                                                                                   & flat tire  response                  & fast adaptation                   \\ \hline
\multirow{2}{*}{\begin{tabular}[c]{@{}l@{}}Education and \\ societal impacts\end{tabular}}                         & driver licence                       & AI ethics and value alignment                        \\ \cline{2-3} 
                                                                                                                   & sustainability                       & green and righteous AI                          \\ \bottomrule
\end{tabular}
\end{adjustbox}
\end{table*}

\section{Analogies between Car and AI Maintenance}
As AI-empowered algorithms and systems are often perceived as a powerful yet mysterious technology to end users, we believe making 
analogies to (autonomous) cars can deliver better transparency and a more comprehensive understanding of AI technology's utilities and limitations. 
Towards formalizing and standardizing the notion of AI maintenance, we aim to draw connections to a more familiar case -- car maintenance -- as AI and car share many commonalities in model development and deployment. The development of new car models is a resource-intensive process (e.g., electric cars). It is taken for granted that
essential regulatory and law requirements such as reliability and safety are fully certified throughout the development process, to avoid catastrophic failures, fatal damage, and critical product recalls.
Similarly,   AI model development can be quite expensive, especially when it comes to the training of foundation models \cite{bommasani2021opportunities} that require pre-training on large-scale datasets with neural networks consisting of a massive number of trainable parameters. Take the Generative Pre-trained Transformer 3 (GPT-3) \cite{brown2020language}  as an example, which is one of the largest language models ever trained to date. GPT-3 has 175 billion parameters and is trained on a dataset consisting of 499 billion tokens. The estimated training cost is about 4.6 million US dollars even with the lowest priced GPU cloud on the market in 2020\footnote{\url{https://lambdalabs.com/blog/demystifying-gpt-3/}}. Having invested so much, one would expect the resulting AI model is risk-proof and robust to be deployed.

In deployment, car maintenance involves regular mechanical and electrical inspection, performance testing and certification, automobile part replacement, and repair. We argue that many familiar concepts in car maintenance can be well-mapped to AI models.
In what follows, we make 
analogies between car and AI to facilitate the consolidation of AI maintenance for robustness.

\textbf{Table \ref{tab: car_AI}} summarizes the key terms that share analogies between car and AI maintenance for robustness. In what follows, we divide those terms into four categories and discuss their connections.

\subsection{Model descriptions and performance characterization}
The ``user manual'' provides instructions for an AI system, with descriptions specifying necessary information for transparency and accountability, such as data and model training details, privacy, usability, and impact statements regarding recommended uses and possible misuse. The ``automobile parts'' in AI means functional and configurable modules in the machine learning pipeline that can be modified and ideally standardized for the ease of model fix and update. The ``warrant'' in AI means qualitative and quantitative performance checkpoints in the development process. The ``transmission efficiency'' in AI relates to how the model scales with data, memory, and power, such as  floating-point operations per second (FLOPS).

\begin{figure*}[t]
    \centering
    \includegraphics[width=0.95\textwidth]{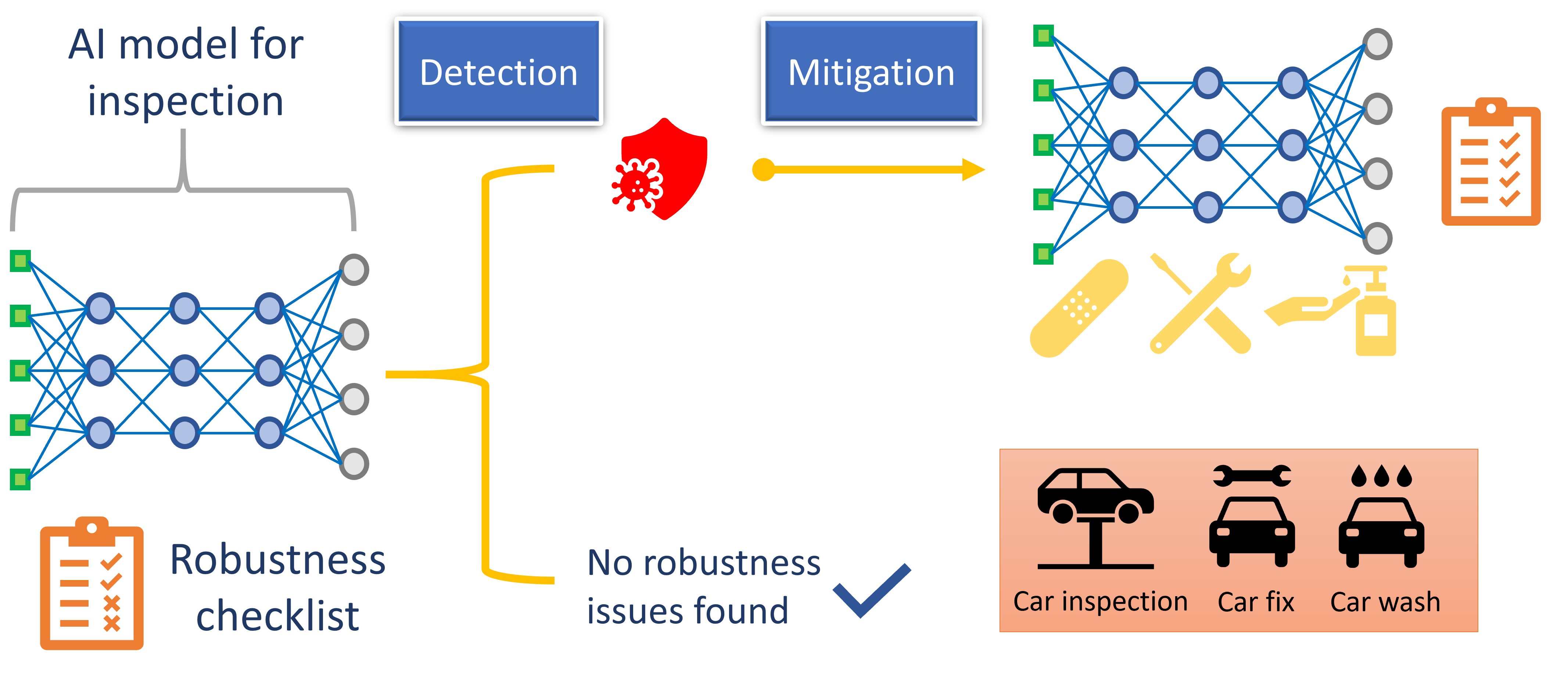}
    \caption{The \textit{AI model inspector} framework consists of detection and mitigation stages. 
    The model under inspection first takes a series of robustness testing and checkpoints, including procedural and operational assessment, passive evaluation on representative datasets, and active probing by generating new instances  on-the-fly to find failure modes. In the detection stage, the inspector extracts statistics and runs a diagnosis to identify possible risks in robustness. In the mitigation stage, the inspector employs model fix and update to mitigate the identified robustness risks, and then re-assesses the model using the same robustness checklist. Finally, the inspector returns a risk-mitigated model.
    The entire process is analog to car inspection, fixing, and cleaning for car maintenance.}
    \label{fig:inspector}
\end{figure*}

\subsection{Systematic inspection and monitoring}
During model development, the ``collision test'' for AI refers to internal comprehensive robustness assessment, white-hat hacking, and red-teaming to identify limitations and hidden issues, similar to comprehensive road testing and car reviews. The results can be used to generate a ``safety report'' providing a quantified level of robustness in adversarial and unseen scenarios. The ``mechanical and electrical inspection'' for AI means penetration testing and debugging of the entire system (e.g., the software and hardware supporting AI technology) using probing and active measurement. The ``problematic status warning'' refers to real-time operational abnormal event detection during deployment, such as erroneous instances or malfunctioning. The ``health state monitoring'' means continuous tracking of model behaviors, such as identifying the emergence of adversarial threats and data drifts.

\subsection{Fix and update}
After inspecting and identifying errors and risks, the ``repair'' for AI models refers to mitigation strategies to fix, update, and re-certify the underlying model. The ``wheel-alignment'' for AI means model calibration, the ``winter tire'' means hardening the model with a more robust module, and the ``flat tire response'' means fast adaption of an AI model in the face of model performance degradation and anomalous events. Depending on the severity of the found robustness risks, user demand, and enforced regulation for AI technology, model fix and update for AI maintenance can have differentiated services at varying costs, ranging from simple model patching and quick problem fixing, module replacement, partial model upgrade, to model rebuild.

\subsection{Education and societal impacts}
The ``driver license'' for AI means education on the ethics  and value alignment when using  AI technology, to understand its capabilities and limitations.
The ``sustainability'' for AI involves gaining environmental awareness such as greener AI models with reduced energy consumption, as well as achieving positive societal impacts, in order to fulfill social responsibility and prevent possible misuse.

\section{AI Model Inspector}

Towards practicing and realizing the notion of AI maintenance, in this section we propose a methodology called AI model inspector, which is a conceptual pipeline for proactive detection and mitigation of robustness issues throughout the AI lifecycle. We also highlight two 
case studies on different robustness challenges to illustrate how the AI model inspector can be realized. Finally, as motivated by vehicle autonomy, we define different levels of AI robustness.

\subsection{Robustness inspection: detection and mitigation}

{\textbf{Figure \ref{fig:inspector} } shows the pipeline of AI model inspector consisting of two stages: \textit{detection} and \textit{mitigation}. First, a user using the AI maintenance service provides a model and/or some data samples for robustness inspection. The inspection takes a series of robustness testing and checkpoints in both qualitative and quantitative manners, including procedural and operational assessment, passive model performance evaluation on representative datasets, and active probing by generating new instances on-the-fly to find failure modes. Qualitative assessment includes soliciting system characterization and problem descriptions from the model operator to gain a comprehensive understanding of the scope and details of model development and deployment, such as what model and data are used for training, how the model is deployed, how much information is known to a user, what types of robustness challenges are of top concerns, to name a few.  Based on the qualitative assessment, quantitative analysis includes running the corresponding diagnosis and reporting the numerical results and summary by generating and leveraging proper test cases and datasets for performance evaluation.

Specifically, in the detection stage, the inspector extracts discriminative statistics and runs a diagnosis to identify possible risks in robustness. Then, in the mitigation stage, the inspector employs model fix and update to mitigate the identified robustness risks, such as model finetuning and re-training, adding or replacing some modules in the AI system, and re-assessing the model using the same robustness checklist. Finally, the inspector returns a risk-mitigated model. The entire process is analog to car maintenance in terms of car inspection, fixing, and cleaning.
The notion of differentiated services in car maintenance can also be mapped to the varying demand and cost of AI maintenance, such as fast scanning, thorough  inspection, quick patching, and detailed fix and update.
We note that the usage of the AI model inspector is continuous rather than one-shot. Based on the recurrence of the states in the AI lifecycle, a model will repeatedly undergo several transitions between the states of data collection \& processing, model training, and model deployment. Moreover, 
a model can be fixed but broken again later. This is analogous to the notion of weariness and fatigue testing in predictive car maintenance  
    -- after inspection, some parts need to be updated or replaced on a regular basis to ensure the model remains in good condition.

Based on the robustness challenges shown in Figure \ref{fig:lifecycle_robustness}, 
we make the following two examples that realize the concept of the AI model inspector.

\begin{table*}[t]
\caption{Comparisons between the levels of vehicle autonomy versus AI robustness. }
\label{tab:level}
\centering
\begin{tabular}{@{}l|l|l@{}}
\toprule
\textbf{Level} & \textbf{Vehicle Autonomy}             & \textbf{AI Robustness}                            \\ \midrule
0     & no driving automation          & no robustness (standard training)        \\
1     & driver assistance               & generalization under distribution shifts \\
2     & partial driving automation     & robustness against single risk           \\
3     & conditional driving automation & robustness against multiple risks        \\
4     & high driving automation        & universal robustness to known risks                     \\
5     & full driving automation        & human-aligned and augmented robustness   \\ \bottomrule
\end{tabular}
\end{table*}

\textit{Backdoor detection and mitigation}: In the detection stage, the inspector adopts the Trojan net detector proposed in  \cite{wang2020practical}, which uses a limited number of untampered clean data samples (as few as one sample per class) to derive a discriminate statistic for discerning a trained neural network classifier has any hidden backdoor. The detector can even achieve data-free detection for convolutional neural networks. After detection, the inspector can adopt the mitigation strategy of model sanitization proposed in
\cite{Zhao2020Bridging} to remove the backdoor by finetuning the model parameters.

\textit{Anomalous input detection and mitigation}: Given a data input to an AI model under inspection, the inspector can 
use internal data representations (e.g., similarity to training data), domain knowledge (e.g., innate data characteristics and physical rules), or external knowledge checking (e.g., searching and reasoning over a knowledge graph or a database) to determine whether the data input is anomalous or not. Here, the anomaly encompasses different robustness challenges, such as adversarial examples and out-of-distribution samples. For instance, the innate temporal dependency in audio data is used in  \cite{yang2018characterizing} to detect audio adversarial examples for automatic speech recognition, and many distance metrics based on the internal data representations extracted from the model have been proposed to detect out-of-distribution samples \cite{yang2021generalized}. In addition to filtering out anomalous inputs, the inspector can further take mitigation strategies to update the model and strengthen its robustness against anomalous inputs. For instance, the self-progressing robust training method proposed in \cite{cheng2020self} can further strengthen a trained model for enhanced adversarial robustness by instructing the model to mitigate the self-discovered ambiguity during model finetuning.

\subsection{Adversarial Machine Learning for Robustness}
    Cars like the Mars Exploration Rovers can successfully execute the assigned task on new and unseen terrain because they were developed in comprehensive simulated environments. For AI models, one can incorporate the failure examples generated from model inspection tools to improve the robustness in unseen and even adversarial environments. This methodology is known as \textit{adversarial machine learning}, by introducing a virtual adversary in the AI lifecycle to help create better and more robust models. In the development phase, the role of the virtual adversary is to simulate the out-of-distribution or worst-case scenarios and generate new challenging cases to help the model generalize better in unseen and adversarial environments. In the deployment case, the role of the virtual adversary is to employ proactive robustness evaluation and risk discovery, in order to prevent real damage and negative impacts. 
    One typical example is \textit{adversarial training} \cite{madry2017towards} which exploits self-generated adversarial examples during model training to strengthen adversarial robustness against adversarial inputs in the deployment phase.
    We refer the readers to \cite{AdvML} for recent advances in adversarial machine learning for AI robustness.

\subsection{Roadmap towards the levels of AI robustness}

Inspired by the definitions for six levels of driving automation for autonomous vehicles\footnote{\url{https://www.sae.org/standards/content/j3016_202104/}}, we define six levels of AI robustness to facilitate technical progress tracing, risk quantification, and inspection, model auditing, and standardization.
{\textbf{Table \ref{tab:level}}} compares the defined levels for vehicle autonomy and AI robustness, respectively. The level of robustness quantifies the progress in the soundness of machine intelligence for robustness.
As the level increases, it signifies the practice and guarantee of robustness in a more practical and comprehensive manner. 
For AI robustness, an increased level means broader coverage of robustness risks under consideration. We believe formalizing the levels of AI robustness can be useful for the discussion and practice of AI standardization related to robustness, security, and safety, such as ISO/IEC JTC 1/WG 13 on Trustworthiness\footnote{\url{https://www.iso.org/committee/45020.html}} and 
ISO/TC 22/SC 32/WG 14 on Safety and Artificial Intelligence\footnote{\url{https://standards.iteh.ai/catalog/tc/iso/6ec701ad-7678-442d-b186-a84b9ba2bbdf/iso-tc-22-sc-32}}.

Level 0 means the original robustness obtained from a standard model training process without any risk mitigation operations. Level 1 concerns the generalization capability on naturally occurring shifted data distributions, such as maintaining robust predictions against distributional changes caused by spurious features that are irrelevant to the actual semantic context (e.g., classifying traffic signs with altering sky backgrounds). Level 2 considers the worst-case robustness against single risk (e.g., adversarial examples), and Level 3 extends to multiple risks, such as the multi-objective (but selected) robustness to adversarial examples, common data corruptions, and spurious correlations \cite{paul2022vision}. Level 4 guarantees universal robustness to all known risks. Here universal robustness means joint effectiveness on all known robustness risks.
Finally, Level 5 aligns robustness with human-centered values and user feedback, and it has the capability to automatically augment new robustness that is complimentary to existing robustness requirements. 
Depending on the requirements (e.g., law regulation) and contexts of the applications, different levels can be necessitated as pre-requisite before deployment. For example, some high-risk AI applications should pass the criterion of higher levels -- similar to the necessary requirements for different driving automation conditions (e.g., driving on highways versus urban environments).  

It is worth noting that the assessment of level-1 robustness can likely be accomplished by static evaluation on a representative dataset or benchmark. However, moving forward to level 2 and above, the validation of worst-case robustness performance also requires model intervention, such as active model scanning and probing for finding failure cases. Moreover, AI model inspector takes proactive steps for detecting and mitigating potential robustness risks, which differs from
existing frameworks such as Factsheets \cite{arnold2019factsheets}, Model Cards  \cite{mitchell2019model}, and Datasheets for Datasets \cite{gebru2021datasheets} that only employ passive model characterization and specification. Finally, in addition to maintenance for AI, one can also adopt AI to improve maintenance, such as predictive maintenance that takes preventive care to AI models based on historical records and risk forecasting.



\section{Concluding Remarks}
This article discusses a novel maintenance framework for robustness in AI technology based on analogies to the development and deployment of car models. To instill and improve trustworthiness in the AI lifecycle, we propose an automated and scalable solution based on the principle of AI model inspector for detecting and mitigating potential risks when lacking robustness. Inspired by vehicle autonomy, we also define different AI robustness levels for formalizing, evaluating, standardizing, and regulating risk-proof AI models. As AI technology is transforming our life, society, and environment with greater width and depth and at a faster speed than cars, we believe the quest for AI maintenance is imminent and necessary.
Beyond robustness, the AI model inspector framework can also be extended to incorporate other dimensions of trustworthy AI,  such as fairness, explainability, privacy, accountability, and uncertainty quantification.


\bibliography{adversarial_learning}
\bibliographystyle{IEEEtran}

\end{document}